\documentclass{article}
\usepackage{graphicx} % more modern
\usepackage{subfigure}

\usepackage{natbib}

\usepackage{algorithm}
\usepackage{algorithmic}

\usepackage{amsmath}
\usepackage{amssymb}
\usepackage{hyperref}

\usepackage[accepted]{icml2013}

\icmltitlerunning{A Powerful Genetic Algorithm for Traveling Salesman Problem}

\begin{document} 
\twocolumn[
\icmltitle{A Powerful Genetic Algorithm for Traveling Salesman Problem}

% It is OKAY to include author information, even for blind
% submissions: the style file will automatically remove it for you
% unless you've provided the [accepted] option to the icml2013
% package.
\icmlauthor{Shujia Liu}{liushujia@yahoo.com}
\icmladdress{Department of Computer Science, Sun Yat-sen University,
            Guangzhou 510006 China}
%\icmlauthor{Your CoAuthor's Name}{email@coauthordomain.edu}
%\icmladdress{Their Fantastic Institute,
%            27182 Exp St., Toronto, ON M6H 2T1 CANADA}

% You may provide any keywords that you 
% find helpful for describing your paper; these are used to populate 
% the "keywords" metadata in the PDF but will not be shown in the document
\icmlkeywords{boring formatting information, machine learning, ICML}

\vskip 0.3in
]

\begin{abstract} 
This paper presents a powerful genetic algorithm (GA) to solve the traveling salesman problem (TSP). To construct a powerful GA, I use edge swapping(ES) with a local search procedure to determine good combinations of building blocks of parent solutions for generating even better offspring solutions. Experimental results on well studied TSP benchmarks demonstrate that the proposed GA is competitive in finding very high quality solutions on instances with up to 16,862 cities.
\end{abstract}
\section{Introduction}
\label{submission}
The traveling salesman problem (TSP) is one of the most cited NP-hard combinational optimization problems because it is so easy to understand but difficult to solve. It is a challenging problem of significant academic value as it is often used as a benchmark problem when new solution approaches are developed.

The simplest heuristic approach for the TSP would be a greedy local search with the k-opt neighbourhood, which is defined as a set of solutions that are transformed from a current loop by replacing at most k edges to construct possible loops. The solution quality will improve with increasing k when the neighbourhood is completely searched at each iteration, but at the cost of rapidly increasing computation time.

This paper presents a sophisticated version of ES, describe the enhancements and provides more instructive analysis of the impact of these enhancements from a new perspective. Moreover, a wider class of benchmarks and instances with up to 15,000 cities are now also considered. The program code of the proposed GA is available in the online supplement of this paper (available at \textbf{\texttt{https://github.com/sugia/GA-for-TSP}}).

Experimental results on 10 well studied TSP benchmark instances (the largest size is 16,862) show that the proposed GA outperforms the state-of-the-art LK-based algorithms in finding very high quality solutions. the GA find optimal or best known solutions for most benchmark instances in a reasonable computation time.

%The remainder of this paper is organized as follows. In section 2, I describe the basic framework of the proposed GA. In section 3, I present the local version of ES and the global version of ES. Experiment results are presented in section 4. Conclusion is provided in section 5.

\section{Basic Framework}
In this section, I describe the basic ideas and outline the proposed GA in this paper.
\subsection{Basic Ideas}
GAs for the TSP usually require longer computation times than efficiently implemented local search based algorithms in order to exercise their capabilities. One reason for this is the nature of the population based search. However, the major reason arises from the fact that crossover operators require more computational cost to generate an offspring solution than do local search operators to evaluate a solution in the neighbourhood.

To reduce the computational cost of ES, I proposed localization of ES with an efficient implementation of this edge swapping operator. A localized version of ES generates, denoted as $P_A$, by replacing relatively few edges with edges selected from the other parent, denoted as $P_B$. This approach enables generation of an offspring solution in less than O(N) time by making use of the fact that it will be similar to $P_A$. In addition, localization of ES contributes to maintaining population diversity coupled with an appropriate GA framework where only $P_A$ is replaced with an offspring solution in the selection for survival. I therefore use only a localized version of ES from the start of the search until it can no longer effectively generate offspring solutions to improve $P_A$.

When the localized version of ES cannot generate an offspring solution that improves $P_A$, the whole genetic algorithm will get trapped into a local optimum. In this case, the number of edges replaced by ES should be increased to further improve $P_A$, and the localized version of ES should be switched to a global version of ES, which exchanges more edges than does the localized version of ES. In this paper, I propose a good local version of ES and a great global version of ES with a sophisticated design concept.

\subsection{GA Framework}
Algorithm 1 gives the basic framework of GA. The population consists of $N_{pop}$ solutions, where $N_{pop}$ is a parameter. Individuals in the population are generated by an appropriate procedure. Here, I use a greedy local search algorithm with the 2-opt neighbourhood, because it is reasonable to use a simple local search procedure to efficiently obtain $N_{pop}$ solutions with a certain level of quality.

The search process of the GA consists of two stages. First, I use a localized version of ES as the crossover operator from the start of the search until no improvement in the best solution is found over a period of generations. After that, I switch to a global version of ES and use it until the end of the search. More precisely, let $G$ be the number of generations, and if the value of $G$ has already been determined and the best solution does not improve over the last $G$ generations, I terminate the local version of ES and proceed to the global version of ES. The global version of ES is also terminated by the same condition, where $G$ is initialized at the beginning of the algorithm.

Because a localized version of ES generates offspring solutions similar to $P_A$, it is reasonable to replace only parent $P_A$, rather than both parents, in order to better maintain population diversity. Here, the offspring solution that replaces parent $P_A$ is selected according to a given evaluation function. The most straightforward evaluation function would be the tour length, but I employ an alternative evaluation function in order to maintain population diversity in a positive manner.

\begin{algorithm}[tb]
	\caption{Genetic Algorithm}
	\label{alg:example}
\begin{algorithmic}
	\STATE Initialize a population $[x_1,\cdots,x_{N_{pop}}]$
	\REPEAT
	\STATE $r(.)$ = a random permutation of $[1,\cdots N_{pop}]$
	\FOR{$i=1$ {\bfseries to} $N_{pop}$}
	\STATE $P_A = x_{r(i)}$
	\STATE $P_B = x_{r(i+1)}$
	\STATE $[c_1,\cdots,c_{N_{ch}}] = ES(P_A,P_B)$
	\STATE $x_{r(i)} = BEST(c_1,\cdots,c_{N_{ch}},P_A)$
	\ENDFOR
	\UNTIL{termination condition is satisfied}
	\STATE {\bfseries return} the best individual solution
\end{algorithmic}

\end{algorithm}

\section{ES Algorithm}
I first present the ES algorithm along with the description of the framework of ES. The local version of ES is described in section 3.2, and the global version of ES is presented in section 3.3. Details of the efficient implementation techniques for the localized version of ES are presented in the online supplement of this paper.

\subsection{ES Framework}
{\bf Step 1}. Let $M_{AB}$ be the undirected graph merged by $P_A$ and $P_B$, defined as $M_{AB}=(V, E_A\cup E_B)$, where $E_A$ is the edge set of $P_A$ and $E_B$ is the edge set of $P_B$.

{\bf Step 2}. Partition all edges of $M_{AB}$ into merged rings(M-rings), where a M-ring is defined as a ring in $M_{AB}$, such that edges of $E_A$ and edges of $E_B$ are alternately linked. 

The partition of the edges into M-rings is always possible, because for any vertex in $M_{AB}$ the number of incident edges of $E_A$ is equal to that of $E_B$. However, the partition is not uniquely determined and I partition the edges randomly into M-rings in the following way. The procedure is started by randomly selecting a vertex. Starting from the selected vertex, trace the edges of $E_A$ and $E_B$ in $M_{AB}$ in turn until an M-ring is found in the traced path, where the edge to be traced next is randomly selected (if two candidates exist) and the traced edges are immediately removed from $M_{AB}$. If a M-ring is found in the traced path (a portion of the traced path including the end may form a M-ring), store it and remove the edges constituting it from the traced path. If the current traced path is not empty, start the tracing process again from the end of the current traced path. Otherwise, start the tracing process by randomly selecting a vertex from among those linked by at least one edge in $M_{AB}$. If there is no edge in $M_{AB}$, iterations of the tracing process are terminated.

{\bf Step 3}. Construct a rings set(R-set) by selecting M-rings according to a given selection strategy, where an R-set is defined as the union of M-rings. Note that this selection strategy determines a version of ES.

{\bf Step 4}. Generate an intermediate solution from $P_A$ by removing the edges of $E_A$ and adding the edges of $E_B$ in the R-set, i.e., generate an intermediate solution by $E_C=(R-set\cap E_A)\cup(R-set\cap E_B)$. An intermediate solution consists of one or more loops.

{\bf Step 5}. Generate an offspring solution by connecting all loops into one loop .

{\bf Step 6}. If a further offspring solution is generated, then go to Step 3. Otherwise, terminate the procedure.

\begin{figure}[ht]
\includegraphics[width=2.9in]{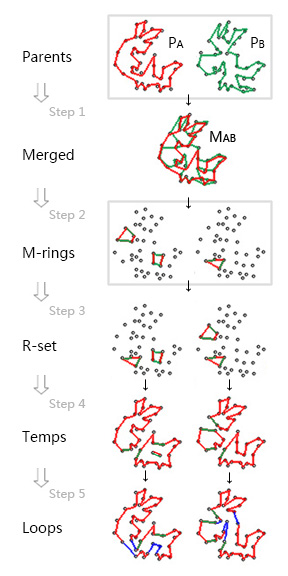}
\caption{Illustration of the Edge Swapping Algorithm}
\end{figure}

I define the size of a M-ring as the number of edges of $E_A$ (or $E_B$) included in it. Note that some of the M-rings might consist of two overlapping edges, one from $E_A$ and one from $E_B$. I call such a M-ring ineffective because the inclusion of ineffective M-rings in an R-set does not affect the resulting intermediate solution. I call an M-ring effective if it includes more than four edges. In Step 3, I select only effective M-rings for constructing R-set. I define the size of a R-set as the number of edges of $E_A$ (or $E_B$) included in it.

According to the definition of a R-set and the procedure in Step 4, ES generates an intermediate solution from $E_A$ by replacing edges with the same number of edges selected from $E_B$, under the condition that every vertex is linked by just two edges. An intermediate solution therefore consists of one or more loops.

\subsection{Local Version of ES}
ES can generate various intermediate solutions, depending on the combination of M-rings selected in Step 3 for constructing R-sets. I can construct different versions of ES by using different selection strategies of M-rings. For the original ES, I proposed a simple selection strategy, which I call single strategy in this paper. In addition, I proposed another simple selection strategy, which is called the random strategy in this paper. The two selection strategies of M-rings are described below.

{\bf Single strategy}. Select a single M-ring randomly without overlapping the previous selections.

{\bf Random strategy}. Select M-rings randomly with a probability of 0.5 for each.

The intermediate solution of the single strategy tends to be similar to $P_A$. In contrast, the random strategy typically forms a R-set in Figure 1, and the resulting intermediate solution tends to contain edges of $E_A$ and edges of $E_B$ equally. In this paper, I apply the random strategy to the construction of a localized version of ES.

\subsection{Global Version of ES}
In this subsection, I present three selection strategies of M-rings for constructing global versions of ES. Here, the size of R-sets generated by a global version of ES should be greater than that by the localized version of ES. However, the increase of the R-set size typically increases the number of sub-loops in intermediate solutions (see the R-sets and the resulting intermediate solutions in Figure 1 for an illustration), which degrades the capability to generate good offspring solutions that improve parent solutions when parent solutions are very high quality loops. This is because the new edges introduced in Step 5 of the ES algorithm frequently degrade the quality of offspring solutions in this situation. So it is preferable to decrease the number of sub-loops while increasing the R-set size. A global version of ES needs to be designed to satisfy these two competing demands.

{\bf K-multiple strategy}. Select K M-rings randomly, where K is a given parameter (K=6 in my experiments).

In this paper, I propose a heuristic selection strategy of M-rings, called block strategy to construct an effective global version of ES. 

{\bf Block strategy}. Select geographically close M-rings to construct a R-set, the resulting intermediate solution is generated from $E_A$ by replacing a block of edges of $E_A$ with a block of edges of $E_B$ in the same region. 

In fact, the block strategy is clearly superior to the K-multiple strategy in the experiments.

\section{Computational Experiments}
The proposed GA was tested on 10 instances with up to 16,862 cities selected from well-known, widely used benchmark sets for the TSP.

They are the TSPLIB:

(\textbf{\texttt{http://comopt.ifi.uni-heidelberg.de}})

and the National TSP benchmarks:

(\textbf{\texttt{http://www.math.uwaterloo.ca/tsp/}}).

 The number of cities for every instance is indicated by the instance name.

The proposed GA was implemented in C and the program code was compiled using GNU gcc compiler. The program code is available in the online supplement. I executed the GA in a computer with Intel(R) Core(TM) i5 CPU 2.27 GHz, and the program execution time varied substantially, depending on the instances. So I measured the CPU time by executing the GA $10$ trials and calculate the average time cost of them to obtain the results.

In this section, I first describe several configurations for the GA and then analyze the impact of the proposed enhancements on a selected group of the instances. Detailed results of the GA using all enhancements on all 10 instances are compared with a LKH algorithm.

\subsection{Configuration of the GA}
I apply the GA (Algorithm 1) using several different configurations to analyze the impact of the proposed enhancements on the performance. One configuration is determined by selecting one strategy from each of the items listed below. For each item, the default strategy corresponds to each of the proposed enhancements, and other strategies are also tested for comparison. As for the default parameter values for $N_{pop}$ and $N_{ch}$, I determined them through preliminary experiments.

Population size ($N_{pop}$): $200$(default). Alternatively, set to $400$ if greedy selection is used.

Number of offspring solutions ($N_{ch}$): $20$(default). Alternatively, set to $10$, $30$, and $40$.

\subsection{Impact of the ES}
I apply the GA using each configuration 10 times to a selected group of instances with sizes ranging from 9,847 to 16,862 in order to save space and to avoid numerous experiments. Results are presented in the following format: the instance name(Instance), the number of runs that succeed in finding the optimal or best known solution (no better solution was found) over 10 runs, the average percentage excess with respect to the optimal solutions(Err), and the computation time for a single run in seconds(Time).

\begin{table}[tb]
{\small
\begin{tabular}{lrlrlr}
\hline
\abovespace
(10 runs)&& LKH &  & GA & \\
Instance & Optimum & Err & Time & Err & Time \\ 
\hline
\abovespace
ja9847   & $491,924$    & $0.07$ & $381$    & $0.00$ & $372$    \\
xmc10150 & $28,387$     & $0.03$ & $403$    & $0.00$ & $395$    \\
rl11849  & $923,288$    & $0.04$ & $549$    & $0.00$ & $451$    \\
xvb13584 & $37,083$     & $0.04$ & $691$    & $0.01$ & $592$    \\
brd14051 & $469,385$    & $0.01$ & $744$    & $0.00$ & $651$    \\
xrb14233 & $45,462$     & $0.02$ & $759$    & $0.01$ & $676$    \\
fnl4461  & $182,566$    & $0.01$ & $785$    & $0.00$ & $705$    \\
rl5915   & $565,530$    & $0.02$ & $882$    & $0.00$ & $799$    \\
rl5934   & $556,045$    & $0.05$ & $928$    & $0.02$ & $815$    \\
it16862  & $557,274$    & $0.01$ & $976$    & $0.00$ & $863$    \\
\hline
\end{tabular}
}
\caption{Results of GA Compared with LKH}
\end{table}

\section{Conclusion}
In this paper I have proposed a powerful GA in finding very high quality solutions for the TSP. The proposed GA has found optimal or best known solutions for most benchmark instances with up to 16,862 cities. One of the strengths of my GA is the use of ES, an edge swapping operator for the TSP. the local version of ES and the global version of ES significantly reduce the computational cost, with the help of efficient implementation techniques. This resolves the common problem that GA for TSP are usually much more time consuming than efficiently implemented local search based algorithms. Another important contribution is the development of ES in generating even better offspring solutions from very high quality parent solutions at the final phase of the GA. An interesting feature is that I design a simple local search procedure into ES to determine good combinations of the edges of parents. I have demonstrated that the enhancements significantly improve the performance of the GA. I believe that the proposed GA provides a good example of a sophisticated GA application for a representative combinatorial optimization problem and that some of the ideas can be successfully applied to the design of GAs for other combinatorial optimization problems.

% In the unusual situation where you want a paper to appear in the
% references without citing it in the main text, use \nocite
%\nocite{langley00}
%\nocite{g0}
\nocite{g1}
%\nocite{g2}
%\nocite{g3}
%\nocite{g4}
%\nocite{g5}
%\nocite{g6}
%\nocite{g7}
%\nocite{g8}
%\nocite{g9}
%\nocite{g10}
%\nocite{g11}
%\nocite{g12}
%\nocite{g13}
%\nocite{g14}
%\nocite{g15}
%\nocite{g16}
%\nocite{g17}
%\nocite{g18}
%\nocite{g19}
%\nocite{g20}
%\nocite{g21}
%\nocite{g22}
%\nocite{g23}
%\nocite{g24}
%\nocite{g25}

\nocite{g26}
\nocite{g27}
\nocite{g28}

\nocite{g30}
\nocite{g31}
\nocite{g32}
\nocite{g33}
\nocite{g34}
\nocite{g35}

\nocite{g36}
\nocite{g37}
\nocite{g38}
\nocite{g39}
\nocite{g40}
\nocite{g41}
\nocite{g42}
\nocite{g43}
\nocite{g44}
\nocite{g45}
\bibliography{paper}

\bibliographystyle{icml2013}

\end{document}